%% file: Template.tex
\documentclass{article}
\usepackage{spconf,amsmath,amsfonts,graphicx,hyperref}
\usepackage{algorithm,algorithmicx,algpseudocode }
\usepackage{multirow}
\usepackage{booktabs}
\usepackage{makecell}

\title{Universal Adversarial Suffixes for Language Models \\Using Reinforcement Learning with Calibrated Reward}
\name{
    Sampriti~Soor\textsuperscript{1}, Suklav~Ghosh\textsuperscript{2}, Arijit~Sur\textsuperscript{2}
    \thanks{
        e-mail ids: SS - \texttt{sampreetiworkid@gmail.com}, \newline SG - \texttt{suklav@iitg.ac.in}, AS - \texttt{arijit@iitg.ac.in}
        \newline Codes: \texttt{\scriptsize https://github.com/SampritiSoor/RLforAdversarialSuffix} 
    }
}
\address{
    \textsuperscript{1}Center for Intelligent Cyber Physical Systems, Indian Institute of Technology Guwahati, India
    \\
    \textsuperscript{2}Department of Computer Science and Engineering, Indian Institute of Technology Guwahati, India
}
\begin{document}
%
\maketitle
\begin{abstract}
Language models are vulnerable to short adversarial suffixes that can reliably alter predictions. Previous works usually find such suffixes with gradient search or rule-based methods, but these are brittle and often tied to a single task or model. In this paper, a reinforcement learning framework is used where the suffix is treated as a policy and trained with Proximal Policy Optimization against a frozen model as a reward oracle. Rewards are shaped using calibrated cross-entropy, removing label bias and aggregating across surface forms to improve transferability. The proposed method is evaluated on five diverse NLP benchmark datasets, covering sentiment, natural language inference, paraphrase, and commonsense reasoning, using three distinct language models: Qwen2-1.5B Instruct, TinyLlama-1.1B Chat, and Phi-1.5. Results show that RL-trained suffixes consistently degrade accuracy and transfer more effectively across tasks and models than previous adversarial triggers of similar genres.
\end{abstract}
\begin{keywords}
Adversarial Suffixes, Reinforcement Learning, Calibrated Reward, Prompt-Based Attacks
\end{keywords}
%
%
%
\section{Introduction}
\label{sec:intro}

Language models (LMs) have rapidly advanced in recent years, showing strong capabilities across a wide range of natural language understanding and generation tasks \cite{radford2019language}. Despite these successes, their vulnerability to carefully crafted adversarial inputs remains a pressing concern. Short trigger phrases added to otherwise benign prompts can reliably alter model predictions \cite{rajpurkar2018know}. Such vulnerabilities undermine trust in LMs for real-world applications, especially in safety-critical or high-stakes domains.

Most prior work on adversarial prompts has treated the problem as a form of discrete optimization, relying on gradient-based search or heuristic substitution to identify token sequences that cause misclassification \cite{rajpurkar2018know,wallace2019universal,zou2023universal}. While effective, these methods are often brittle: they exploit local gradient information but do not systematically explore the broader space of candidate suffixes. As a result, the discovered triggers may lack transferability across tasks and models, and even often fail when the evaluation setting changes \cite{liang2022holistic,schick2020exploiting}.

In this work, we take a different perspective by casting adversarial suffix generation as a reinforcement learning (RL) problem. Instead of greedily selecting token substitutions, we treat the suffix as a policy over discrete tokens. The policy samples candidate suffixes, evaluates them using a calibrated reward signal derived from classification cross-entropy, and updates iteratively using policy gradient. This allows adversarial search to balance exploration (trying diverse suffix candidates) and exploitation (reinforcing suffixes that systematically degrade performance). Crucially, the base language model remains frozen; only the suffix policy is optimized, making the approach lightweight and efficient.

\noindent Our main contributions are:

\noindent{\bf RL-based adversarial suffixes:} We frame adversarial suffix generation as a reinforcement learning problem, using policy gradient instead of greedy or gradient substitution.  

\noindent{\bf Calibrated reward:} We design the reward with contextual calibration \cite{zhao2021calibrate}, making suffixes act beyond label-surface bias and remain effective in zero-shot evaluation.  

\noindent{\bf In-domain evaluation:} We validate on five diverse tasks showing that RL suffixes create stronger accuracy shifts and calibrated effects than earlier trigger methods.

%
%
\input{tables/relatedWorks_new}

%
%
\section{Methodology}

\newcommand{\wrap}{\operatorname{wrap}}
\newcommand{\CE}{\mathrm{CE}}
\newcommand{\CalCE}{\mathrm{CalCE}}
\newcommand{\Agg}{\mathrm{Agg}}
\newcommand{\softmax}{\mathrm{softmax}}

Here, we formalize adversarial suffix generation as a reinforcement learning problem over a frozen language model. 
Let $\mathcal{M}$ denote the \emph{seen model}, with input vocabulary $\mathcal{V}$ and embedding dimension $H$. 
We assumed $\mathcal{M}$ is frozen, which means it serves as a black-box reward oracle without parameter updates. 
Our objective is to learn a short suffix $s = (t_1, \dots, t_K)$, where each $t_i \in \mathcal{V}$, that when appended to arbitrary prompts systematically alters the model's predictions in a calibrated way. 

Let $(x, y)$ denote a task instance, where $x$ is a natural language input and $y \in \mathcal{Y}$ is the gold label from a discrete label space $\mathcal{Y}$. 
For a wrapped prompt $w(x)$ (including system instructions), an answer prefix $p$, and a suffix $s$, the model defines conditional probabilities
\begin{equation}
P_\mathcal{M}(y \mid w(x), s, p).
\end{equation}
The adversary seeks a suffix $s$ that maximizes an attack objective $J(s)$ aggregated over tasks, prompts, and labels. 
Since suffixes are discrete, we parameterize a stochastic policy $\pi_\theta(s)$ over $\mathcal{V}^K$ and optimize $\theta$ with reinforcement learning. 

\vspace{.25cm}\noindent\textbf{Reward Design:}
The reward signal is based on calibrated cross-entropy ($\CalCE$). 
For each candidate suffix $s$ and task instance $(x,y)$, the calibrated score is defined using context cross-entropy $\CE_{\text{ctx}}$ and null cross-entropy $\CE_{\text{null}}$ as
\begin{align}
\CE_{\text{ctx}}(s; x,y) &= -\log P_\mathcal{M}(y \mid w(x), s, p), 
\\
\CE_{\text{null}}(y) &= -\log P_\mathcal{M}(y \mid p),
\\
\label{eq:calce}\CalCE(s; x,y) &= \CE_{\text{ctx}}(s; x,y) \;-\; \CE_{\text{null}}(y).
\end{align}
By subtracting the null-prompt baseline, the reward focuses only on the effect of the actual input and any modification introduced by the adversarial suffix, and ensures that predictions are not dominated by the model’s inherent bias toward certain label tokens.

To handle variation in label surface forms, the calibrated scores are further aggregated across all surface realizations $\sigma(y) \subset \mathcal{V}^*$ of the same label using a log-sum-exp (lse) operation. 
\begin{equation}
R_{\text{cal}}(s; x,y) =  -\log \sum_{\tilde{y} \in \sigma(y)} 
    \exp\!\big(-\CalCE(s; x, \tilde{y})\big)
\end{equation}
This soft-min aggregation favors suffixes that consistently increase error across different label tokens, rather than exploiting only a single surface form.

Two auxiliary penalties are added to stabilize learning.  
\\
\textbf{(1) Fluency penalty \cite{cheng2020seq2sick}:} Let $\text{CE}_{\text{LM}}(s)$ denote the cross-entropy loss of the suffix $s$ under the frozen model $\mathcal{M}$, i.e.\ how unlikely the sequence is according to $\mathcal{M}$’s own language modeling distribution. The penalty $\lambda_{\text{fl}}\text{CE}_{\text{LM}}(s)$ discourages degenerate or low-probability strings, keeping suffixes plausible.  
\\
\textbf{(2) KL-to-uniform penalty \cite{schulman2017proximal}:} Let $\pi_\theta$ be the suffix policy and $U$ the uniform distribution over the vocabulary. The penalty $\beta D_{\text{KL}}(\pi_\theta \,\Vert\, U)$ prevents premature collapse of $\pi_\theta$ to one-hot distributions, maintaining exploration across tokens.  

Together, these terms regularize training, ensuring the learned suffixes remain both effective and transferable. The complete reward is
\begin{equation}
\hat{R}(s;x,y) = R_{\text{cal}}(s; x,y) - \lambda_{\text{fl}} \,\CE_{\text{LM}}(s) 
- \beta \, D_{\text{KL}}(\pi_\theta \,\Vert\, U)
\end{equation}

Since raw reward estimates can have high variance across prompts, a value baseline is subtracted to stabilize training. For each input 
$x$, we maintain a baseline predictor $b(x)$, and define the advantage estimate as
\begin{equation}
A(s;x)= \hat{R}(s;x)\;-\;b(x)
\end{equation}
This centers rewards around zero, improving the stability of policy-gradient updates. We implement $b(x)$ as a simple moving average over past rewards for each task.

\vspace{.25cm}\noindent\textbf{Policy Parameterization and Optimization with PPO:}
The suffix policy $\pi_\theta(s)$ is factored into independent categorical distributions:
\begin{equation}
\pi_\theta(s) = \prod_{i=1}^K \pi_\theta(t_i \mid i),
\end{equation}
with logits $\theta \in \mathbb{R}^{K \times |\mathcal{V}|}$, meaning each position $t_i$ in the suffix is sampled independently, with its own categorical distribution over the vocabulary. This design gives efficient parallel sampling. 

We optimize $\pi_\theta$ using Proximal Policy Optimization (PPO) \cite{schulman2017proximal}. 
The PPO objective is
\begin{equation}
\mathcal{L}_{\text{PPO}}(\theta) = 
\mathbb{E}_{s \sim \pi_\theta} \left[\frac{\pi_\theta(s)}{\pi_{\theta_{\text{old}}}(s)}
A(s) \right]
\end{equation}

To prevent premature collapse of the suffix policy and to stabilize PPO updates under noisy rewards, we include standard regularizers: an entropy bonus $\eta H[\pi_\theta]$ to encourage exploration \cite{schulman2017proximal} and an optional KL-to-old penalty $\gamma D_{\text{KL}}(\pi_\theta \Vert \pi_{\theta_{\text{old}}})$ to avoid large policy shifts \cite{ziegler2019fine,ouyang2022training}. 
We maximize the PPO objective with respect to the policy parameters. Since standard optimizers perform gradient descent, in practice we minimize the negative of the objective, which is equivalent to performing ascent.
\begin{equation}
\theta = \theta - \alpha \nabla_\theta 
\Big(-\mathcal{L}_{\text{PPO}} 
- \eta H[\pi_\theta] 
+ \gamma D_{\text{KL}}(\pi_\theta \Vert \pi_{\theta_{\text{old}}}) \Big)
\end{equation}
with $\pi_{\theta_{\text{old}}}$ periodically synchronized to the current policy $\pi_\theta$.
Gradient updates are computed on batches of task instances drawn from multiple datasets, reducing variance and improving transfer. 

%
%
\section{Results and Evaluation}
\label{sec:res}


\subsection{Experimental Setup}

\noindent\textbf{Models and Datasets:} 
We evaluate adversarial suffix generation on three representative large language models of different genre and scale: Qwen2-1.5B Instruct (instruction-oriented), Phi-1.5 (compact language understanding backbone), and TinyLlama-1.1B Chat (efficient dialogue model). For each run, one model is treated as the \emph{seen} model used to generate suffixes, while the other two serve as \emph{unseen} models for transfer evaluation.
Five tasks are considered to cover diverse NLP objectives: sentiment analysis (SST-2), natural language inference (RTE), paraphrase detection (MRPC), commonsense QA (BoolQ), and physical reasoning (PIQA).
All training and evaluations are performed on the \emph{training and validation splits} provided by the respective datasets as implemented in \texttt{torchvision}, ensuring consistency and reproducibility across tasks.

\noindent\textbf{Training Specifications:} 
We experiment with different suffix lengths in terms of number of tokens $K \in \{4, 6, 10\}$ to study the effect of token budget. 
We train for $200$ iterations using a rollout sampler that draws batches of size $32$ from multiple tasks per update. 
Task-specific label surfaces are expanded from a base set of lexical variants (e.g., \textit{``yes'', ``Yes'', ``yes.''}, or \textit{``positive'', ``right'', ``correct''}), ensuring that reward calibration accounts for natural response diversity. 
For all tasks, the answer prefix is fixed as 
\(
p = \texttt{"\textbackslash nThe answer is: "}
\)
which consistently anchors the expected label position across models.
Model-specific preprocessing follows the wrapper type, aligning suffix placement with chat-style (\texttt{chatml}) for TinyLlama-1.1B-Chat-v1.0, instruction-style (\texttt{alpaca}) for Qwen2-1.5B-Instruct, or raw classification prompts for phi-1.5. 
Hyperparameters include maximum input length 256, entropy regularization ($10^{-3}$), and a learning rate of $5\times 10^{-2}$. 
Training is stabilized with gradient clipping, NaN/Inf guards, and a temperature floor.

\noindent\textbf{Metrics:} 
Baseline evaluation reports \emph{classification accuracy} (Acc), the fraction of correctly predicted labels, and \emph{calibrated cross-entropy} (CalCE, defined in Eq.~\ref{eq:calce}). 
For each input, CalCE subtracts the null-prompt bias from the context-prompt cross-entropy, so that lower values indicate weaker confidence in the correct label after accounting for priors. 

Say, each label $y \in \mathcal{Y}$ has a set of label surfaces $\mathcal{L}(y)$. For input $x$, we predict label $\hat{y}(x)$ as:
\begin{align}
\ell(y|x) &= \log \sum_{s \in \mathcal{L}(y)} \exp\!\big(\log p(s|x)\big)\\
 \hat{y}(x) &= \arg\max_{y \in \mathcal{Y}} \; \ell(y|x).
\end{align}

In transfer experiments, we report relative changes $\Delta$Acc = Acc\textsubscript{attacked} $-$ Acc\textsubscript{clean} and $\Delta$CalCE = CalCE\textsubscript{attacked} $-$ CalCE\textsubscript{clean}. 
Since CalCE is lower-is-worse, stronger attacks manifest as more negative $\Delta$Acc and more positive $\Delta$CalCE.

\input{tables/baseline}
\input{tables/transferability}

\subsection{Baseline Performance}
Table \ref{tab:baseline} contains zero-shot accuracy and CalCE across tasks and models. Qwen2-1.5B shows the strongest overall performance, combining high accuracy with well-calibrated confidence, consistent with its instruction tuning. Phi-1.5 offers balanced behavior, often outperforming TinyLlama in both accuracy and calibration despite its smaller scale. TinyLlama attains moderate CalCE but struggles to translate this into accuracy, suggesting limited discriminative power. Task-level results show that SST-2 and RTE are relatively well captured by Qwen2-1.5B, while PIQA remains difficult across all models. BoolQ stands out for strong calibration even when accuracy lags, indicating that models can recognize plausible answers but often mis-rank them.

\subsection{Transferability}
Table~\ref{tab:transfer} presents the transferability results across models and tasks. A consistent pattern is observed: adversarial suffixes generally yield negative $\Delta$Acc while producing positive $\Delta$CalCE, indicating that the attack shifts calibrated evidence in the intended direction rather than introducing random noise. When Qwen2-1.5B serves as the seen model, the effect strengthens with longer suffixes: accuracy decreases more sharply (e.g., SST-2) while $\Delta$CalCE rises, and this signal partially carries over to Phi and TinyLlama with smaller magnitudes. BoolQ is distinctive, showing almost unchanged accuracy but sharp $\Delta$CalCE gains, which reflects boundary tilting without frequent label flips, precisely what calibration-aware rewards encourage. A scale-asymmetry is also evident: suffixes trained on Phi and transferred up to Qwen produce strong positive $\Delta$CalCE with only mild accuracy drops, while transfer down to TinyLlama sometimes results in negative $\Delta$CalCE (e.g., SST-2, PIQA), likely due to capacity and prior mismatches. Task-level differences persist: MRPC shows stable harmful transfer with moderate $\Delta$Acc decreases and small $\Delta$CalCE increases, effects on RTE remain weak overall, and PIQA is model-dependent, positive on Qwen but unstable on TinyLlama. Overall, increasing $K$ primarily boosts $\Delta$CalCE, while $\Delta$Acc saturates, consistent with calibrated rather than purely disruptive adversarial behavior.


\input{tables/methodComparison_experiemnt}

\subsection{Comparison with Previous Methods}
We compare our approach with two well-known methods for adversarial prompt generation: Universal Triggers \cite{wallace2019universal} and AutoPrompt \cite{shin2020autoprompt}. These methods were chosen because they represent the most direct line of work on learning short adversarial sequences that can consistently influence model predictions across tasks. Universal Triggers rely on gradient-based updates of discrete tokens, while AutoPrompt uses gradient signals to select token replacements that mimic useful features. 
To ensure a fair and unified evaluation, we adapted all methods to the same experimental setup. Each method was applied to the Qwen2-1.5B model with suffix token length $K=4$, inserted into the same instruction-style wrapper and evaluated under the same calibrated cross-entropy scoring with label surface aggregation. This allows us to compare the relative robustness of different approaches under consistent conditions, even though the original works used slightly different architectures and evaluation metrics.
The results in Table~\ref{tab:modelcomparison} show a clear pattern. Universal Triggers and AutoPrompt create only small changes in accuracy and calibrated scores in this stricter setup, while our proposed RL-based suffix achieves more reliable boundary shifts, particularly on tasks like MRPC and RTE. This demonstrates that calibration-aware reinforcement learning provides stronger and more stable adversarial control compared to earlier gradient-only methods.

%
%
\section{Conclusion}
We proposed a reinforcement learning method for adversarial suffixes, guided by a calibrated reward that avoids label-surface bias. The approach is lightweight yet effective, producing stronger accuracy drops and meaningful CalCE shifts, with promising transferability across different tasks and models, extending beyond prior trigger methods.
%
%

\newpage
\bibliographystyle{IEEEbib}
\bibliography{refs}

\end{document}

%% file: tables/relatedWorks_new.tex
\section{Related Works}
\label{sec:relatedworks}

Reinforcement learning (RL) has been used for text generation when direct supervision is not enough. Early works showed that sequence-to-sequence models can be trained with policy gradient to maximize rewards such as BLEU or accuracy instead of predicting tokens step by step \cite{ranzato2015sequence,bahdanau2016actor}. Proximal Policy Optimization (PPO) \cite{schulman2017proximal} improved stability with a clipped objective and later became the main method in reinforcement learning from human feedback (RLHF) \cite{ziegler2019fine,ouyang2022training}. In RLHF, large models are tuned with a reward from a preference model while staying close to the original model by a KL penalty. Our method also uses PPO, but it is different because we do not tune the full model. We only learn a short suffix policy, and the base model remains frozen.

Work on adversarial prompts is more closely related to what we do. Wallace et al.\ \cite{wallace2019universal} showed that short universal triggers can push models to wrong predictions across tasks. Shin et al.\ \cite{shin2020autoprompt} introduced AutoPrompt, where gradients are used to find trigger tokens that make the model show task knowledge. More recent work \cite{zou2023universal,wei2023jailbroken} explored jailbreak suffixes for aligned chat models, proving that short continuations can bypass safety rules. These methods mainly rely on greedy search or direct gradient tricks. In contrast, we design suffix discovery as an RL problem: tokens are sampled from a policy, scored by a calibrated reward, and updated with policy gradient. This gives a balance between exploration and exploitation that earlier approaches do not provide.

Another important challenge comes from label surfaces. In classification, the same label may appear in many lexical forms which can have very different probabilities, which makes evaluation unstable and easy to attack. Contextual calibration \cite{zhao2021calibrate,holtzman2021surface} reduces this problem by subtracting a null-prompt score, so results reflect only the input context and not the model’s prior bias. This is very important in zero-shot settings, where models often fall back to label priors. Our method includes this calibration inside the reward itself, so suffixes must change the decision boundary in a meaningful way. Compared to prompt tuning \cite{li2021prefix,lester2021power} or RLHF \cite{ziegler2019fine,ouyang2022training}, our method only learns short discrete suffixes, which makes it light and transferable. Compared to trigger or jailbreak attacks \cite{wallace2019universal,wei2023jailbroken}, our calibrated objective makes the adversary more robust across tasks and models.

%% file: tables/baseline.tex

\begin{table}[!b]
  \centering
  \caption{0-shot baseline; each cell reports Acc/CalCE.}
  \label{tab:baseline}
  \begin{tabular}{lccc}
  \toprule
  \textbf{Task} & \textbf{Qwen2-1.5B} & \textbf{Phi-1.5} & \textbf{TinyLlama} \\
  \midrule
  SST-2   & 0.89 / -8.60 & 0.70 / -4.60 & 0.50 / -5.18 \\
  RTE     & 0.83 / -5.19 & 0.52 / -4.53 & 0.53 / -3.25 \\
  MRPC    & 0.74 / -6.11 & 0.70 / -4.25 & 0.70 / -2.67 \\
  BoolQ   & 0.70 / -3.72 & 0.69 / -3.50 & 0.52 / -3.06 \\
  PIQA    & 0.65 / -3.40 & 0.50 / -2.62 & 0.53 / -0.44 \\
  \bottomrule
  \end{tabular}
\end{table}

%% file: tables/transferability.tex
\begin{table*}[t] 
\centering 
\caption{Transferability results; each cell shows $\Delta$Acc/$\Delta$CalCE relative to the corresponding baseline. Rows are grouped by the \emph{seen model} used to train the PPO suffix; columns show evaluation on each target model under 0-shot for $K \in \{4,6,10\}$.}
\label{tab:transfer} 
\resizebox{\linewidth}{!}{ 
\begin{tabular}{clccc|ccc|ccc} 
\toprule 
\multirow{2}{*}{\rotatebox{90}{\makecell{\bf Seen\\\bf Model}}} & \multirow{2}{*}{\textbf{Task}} & \multicolumn{3}{c}{\textbf{Qwen2-1.5B }} & \multicolumn{3}{c}{\textbf{Phi-1.5 }} & \multicolumn{3}{c}{\textbf{TinyLlama }} \\
\cmidrule(lr){3-5} \cmidrule(lr){6-8} \cmidrule(lr){9-11} & & (K=4) & (K=6) & (K=10) & (K=4) & (K=6) & (K=10) & (K=4) & (K=6) & (K=10) \\ 
\midrule 

\multirow{5}{*}{\rotatebox{90}{\textbf{Qwen2-1.5B}}}
 & SST-2 & \,-0.14 / +0.55\, & \,-0.09 / +0.60\, & \,-0.21 / +1.72\,
         & \,-0.31 / +0.12\, & \,-0.31 / +0.16\, & \,-0.34 / +0.43\,
         & \,-0.13 / +0.22\, & \,-0.13 / +0.15\, & \,-0.24 / +0.38\, \\
 & RTE   & \,-0.22 / +0.26\, & \,-0.21 / +0.26\, & \,-0.24 / +0.40\,
         & \,-0.15 / +0.10\, & \,-0.15 / +0.12\, & \,-0.15 / +0.13\,
         & \,-0.07 / +0.23\, & \,-0.12 / +0.33\, & \,-0.10 / +0.21\, \\
 & MRPC  & \,-0.46 / +0.47\, & \,-0.50 / +0.48\, & \,-0.43 / +0.43\,
         & \,-0.46 / -0.22\, & \,-0.46 / -0.03\, & \,-0.46 / -0.28\,
         & \,-0.13 / +0.19\, & \,-0.42 / +0.26\, & \,-0.39 / +0.16\, \\
 & BoolQ & \,-0.11 / +0.43\, & \,-0.14 / +0.43\, & \,-0.19 / +0.86\,
         & \,-0.02 / +0.09\, & \,-0.03 / +0.23\, & \,-0.12 / +0.21\,
         & \,+0.07 / +0.95\, & \,-0.04 / +0.71\, & \,+0.04 / +1.39\, \\
 & PIQA  & \,-0.04 / +0.45\, & \,-0.06 / +0.33\, & \,+0.00 / +0.81\,
         & \,-0.00 / +0.11\, & \,-0.00 / +0.67\, & \,-0.00 / +0.46\,
         & \,-0.03 / +1.36\, & \,-0.03 / +1.14\, & \,-0.02 / +1.48\, \\
\midrule
\multirow{5}{*}{\rotatebox{90}{\textbf{Phi-1.5}}}
 & SST-2 & \,-0.11 / +2.08\, & \,-0.21 / +1.04\, & \,-0.11 / -0.25\,
         & \,-0.30 / +0.36\, & \,-0.31 / +0.77\, & \,-0.31 / +0.10\,
         & \,-0.03 / -0.55\, & \,-0.03 / -0.58\, & \,-0.03 / -0.12\, \\
 & RTE   & \,-0.05 / +0.19\, & \,-0.03 / +0.28\, & \,-0.05 / +0.30\,
         & \,-0.15 / +0.15\, & \,-0.15 / +0.12\, & \,-0.15 / +0.13\,
         & \,+0.00 / +0.06\, & \,+0.00 / +0.34\, & \,-0.10 / +0.19\, \\
 & MRPC  & \,-0.46 / +0.09\, & \,-0.49 / -0.10\, & \,-0.45 / -0.23\,
         & \,-0.46 / +0.06\, & \,-0.46 / +0.02\, & \,-0.46 / +0.07\,
         & \,-0.08 / -0.03\, & \,-0.14 / -0.02\, & \,-0.21 / -0.12\, \\
 & BoolQ & \,+0.00 / +1.94\, & \,-0.06 / +0.55\, & \,-0.06 / -0.79\,
         & \,-0.03 / -0.02\, & \,-0.02 / +0.50\, & \,-0.03 / +0.23\,
         & \,+0.15 / +1.87\, & \,+0.10 / +1.62\, & \,+0.13 / +1.84\, \\
 & PIQA  & \,-0.05 / +2.05\, & \,-0.04 / +0.80\, & \,-0.03 / -0.16\,
         & \,+0.00 / +0.40\, & \,+0.00 / +0.34\, & \,+0.00 / +0.36\,
         & \,-0.03 / -0.16\, & \,-0.03 / -0.16\, & \,-0.05 / -0.15\, \\
\midrule
\multirow{5}{*}{\rotatebox{90}{\textbf{TinyLlama}}}
 & SST-2 & \,-0.14 / +0.22\, & \,-0.07 / +0.47\, & \,-0.13 / +0.46\,
         & \,-0.03 / -0.52\, & \,-0.31 / +0.54\, & \,-0.03 / -0.74\,
         & \,-0.28 / +0.23\, & \,-0.31 / +0.54\, & \,-0.13 / +0.63\, \\
 & RTE   & \,-0.08 / -0.17\, & \,-0.03 / -0.23\, & \,-0.03 / -0.28\,
         & \,-0.14 / -0.55\, & \,-0.03 / -0.31\, & \,-0.07 / -0.42\,
         & \,-0.15 / +0.22\, & \,-0.15 / +0.25\, & \,-0.15 / +0.25\, \\
 & MRPC  & \,-0.50 / +0.23\, & \,-0.47 / +0.28\, & \,-0.46 / +0.29\,
         & \,-0.43 / -0.45\, & \,-0.42 / -0.23\, & \,-0.46 / -0.32\,
         & \,-0.46 / +0.13\, & \,-0.46 / +0.11\, & \,-0.46 / +0.11\, \\
 & BoolQ & \,-0.07 / -1.05\, & \,-0.08 / -0.52\, & \,-0.08 / -0.31\,
         & \,-0.16 / -0.54\, & \,-0.16 / +1.50\, & \,-0.17 / +1.00\,
         & \,-0.03 / +0.19\, & \,-0.05 / +0.55\, & \,-0.15 / +0.75\, \\
 & PIQA  & \,-0.09 / -0.38\, & \,-0.03 / +0.75\, & \,-0.05 / +0.64\,
         & \,-0.03 / -1.61\, & \,-0.03 / -1.26\, & \,-0.03 / -1.37\,
         & \,+0.00 / +0.17\, & \,+0.00 / +0.19\, & \,+0.00 / +0.23\, \\
\bottomrule 
\end{tabular} 
} 
\end{table*}

%% file: tables/methodComparison_experiemnt.tex
\begin{table}[!h]
  \centering
  \caption{Comparison of adversarial suffix methods on Qwen2-1.5B with $K=4$. 
  Each cell shows $\Delta$Acc/$\Delta$CalCE.}
  \label{tab:modelcomparison}
  \resizebox{\linewidth}{!}{
  \begin{tabular}{lccc}
  \toprule
  \textbf{Task} & \textbf{Universal Trigger \cite{wallace2019universal}} & \textbf{AutoPrompt \cite{shin2020autoprompt}} & \textbf{Proposed method} \\
  \midrule
  SST-2   & -0.07 / +1.10 & -0.08 / +1.40 & \textbf{-0.14} / +0.55 \\
  RTE     & -0.15 / +0.20 & -0.03 / +0.25 & \textbf{-0.22} / +0.26 \\
  MRPC    & -0.21 / +0.10 & -0.04 / +0.25 & \textbf{-0.46} / +0.47 \\
  BoolQ   & -0.08 / +0.60 & -0.01 / +0.80 & \textbf{ -0.11} / +0.23 \\
  PIQA    & \textbf{-0.16} / +0.30 & -0.04 / +0.10 & -0.04 / +0.45 \\
  \bottomrule
  \end{tabular}
  }
\end{table}